\documentclass[english]{cccconf}
\usepackage[comma,numbers,square,sort&compress]{natbib}
\usepackage{epstopdf}
\usepackage{graphicx}
\usepackage{tikz}
\usepackage{pgfplots}
\pgfplotsset{compat=1.17}

\usepackage{url}
\usepackage{graphics} 
\usepackage{epsfig} 
\usepackage{mathptmx} 
\usepackage{mathrsfs}
\DeclareMathAlphabet{\mathcal}{OMS}{cmsy}{m}{n}
\usepackage{cuted}

\usepackage{graphicx}
\usepackage{textcomp}

\usepackage{setspace}
\usepackage{amssymb,amsmath,amsfonts}
\usepackage{booktabs}
\usepackage{makecell}

\usetikzlibrary{decorations.text}
\usetikzlibrary{arrows}

\newtheorem{remark}{\textbf{Remark}}
\newtheorem{assumption}{\textbf{Assumption}}

\newcommand{\Zb}{{\mathbb{Z}}}

\newcommand{\m}{\text{-}}

\begin{document}

\title{Aggressive Racecar Drifting Control Using Onboard Cameras and Inertial Measurement Unit}
\author{Shuaibing Lin\aref{shu},
        Jialiang Qu\aref{shu},
        Zishuo Li\aref{thu},
    	Xiaoqiang Ren\aref{shu},
    	Yilin Mo\aref{thu}
}

\affiliation[shu]{School of Mechatronic Engineering and Automation, Shanghai University, Shanghai 200444, P.~R.~China}
\affiliation[thu]{Department of Automation and BRNist, Tsinghua University, Beijing 100084, P.~R.~China}

\maketitle

\begin{abstract}
Complex autonomous driving, such as drifting, requires high-precision and high-frequency pose information to ensure 
  accuracy and safety, which is notably difficult when using only onboard sensors. In this paper, we propose a drift controller with two feedback control loops: 
  sideslip controller that stabilizes the sideslip angle by tuning the front wheel steering angle, and circle controller that maintains a stable trajectory radius and circle center by controlling the wheel rotational speed. 
  We use an extended Kalman filter to estimate the state. A robustified KASA algorithm is further proposed to accurately estimate the 
  parameters of the circle (i.e., the center and radius) that best fits into the current trajectory. 
 On the premise of the uniform circular 
  motion of the vehicle in the process of stable drift, we use angle information instead of acceleration 
  to describe the dynamic of the vehicle. 
  We implement our method on a 1/10 scale race car. The car drifts stably with a given 
  center and radius, which illustrates the effectiveness of our method.
\end{abstract}



\section{Introduction}

Autonomous driving has made a lot of progress. However, controls in high complexity environments, such as 
  drifting, are still in the exploratory stage \cite{drift_yang, goh2020toward, cutler2016autonomous, jelavic2017autonomous, ZHANG20171916}.
  In drift control, to ensure control efficiency and accuracy, the controller needs to run at 
  high frequency, hence the update frequency of vehicle status should also be high enough.

In \cite{drift_yang}, a motion capture system was used for positioning and tracking, with 
  a controller working at 100Hz on a 1/10 race car. It realized a stable drift 
  with a given center and radius; 
  Goh \textit{et al.} \cite{goh2020toward} applied an Oxford Technical Systems RT4003 dual-antenna integrated 
  RTK-GPS/Inertial Measurement Unit (IMU) to obtain vehicle state information at 250Hz with a controller 
  working at 250Hz on a full-size car, which is able to drift along the reference path. 
  Cutler \textit{et al.} \cite{cutler2016autonomous} used onboard sensors to measure turn rate and wheel speed, 
  and motion capture system to obtain body-frame velocities. A small robotic car applied with reinforcement 
  learning drifts along the reference path. 
  These three methods rely on off-board sensors for high-frequency and high-precision vehicle 
  state information. However, off-board sensors are expensive and the working environment needs to meet certain 
  requirements. 

Onboard sensors give the advantage of having a low cost, and increased system autonomy and mobility. 
  Furthermore, onboard sensors are suitable for both outdoor and indoor environments. 
  Jelavic \textit{et al.} \cite{jelavic2017autonomous} used only onboard sensors (i.e. IMU, encoders) for 
  state estimation of 1/10 race car, which can complete a parking action through drift. 
  Zhang \textit{et al.} \cite{ZHANG20171916} used the onboard camera to estimate the longitudinal and lateral velocity 
  of 1/10 race car. The car can complete a U-turn through drift. 
  These two works use only onboard sensors for drift control. However, we notice that the drift in these 
  two works is transient.
  


In this paper, we propose a multi-sensor scheme using onboard cameras and inertial measurement unit (IMU). 
  Using RGB data and depth data from onboard cameras, we leverage a target recognition algorithm to 
  calculate the precise position of the vehicle. 
  At the same time, another camera is used for visual tracking algorithm to make up for the lack of 
  field of view of a single camera at certain position, and hence improve the vehicle status update frequency and meet the control 
  requirements.
  We also specialize in the estimator for the drifting environment, 
  with a simplified set of parameters to characterize the state of the vehicle. That is, 
  angle information instead of acceleration is used to describe the dynamics of the vehicle.
  
The main contributions of this paper are as follows. We propose an onboard sensor estimation scheme 
  supporting the control of a stable circular drifting task with high frequency and accuracy. 
  Only onboard cameras and IMU are used, which is 
  economic and suitable for both indoor and outdoor scenarios. By combining with the tracking algorithm of the ZED camera, the update frequency of 
  vehicle status can reach a maximum of 160Hz, which can ensure the normal operation of the control 
  algorithm at the frequency of 100Hz. The accuracy of our estimation algorithm 
  when the anchor is in the field of view of D435i is comparable to that of the motion capture system. 
  Moverover, our vehicle can drift stably around a given center with a specified radius for a long time.

This paper follows with Section 2, which presents the problem formulation and notations as well as 
  an overview of our proposed control architecture. Our state estimation algorithm is proposed in Section 3. 
  Section 4 presents the experiment result of our algorithm on our race car, and finally, Section 5 concludes the paper.

\section{Problem Formulation and System Scheme}

In this paper, we aim at the designing of an onboard estimation scheme and a control strategy that can 
  achieve vehicle drifting maneuvers tracking circular paths with an assigned center and radius. 
  The circle drifting task requires high accuracy and update frequency of estimators, which is an 
  appropriate benchmark for evaluating the performance of the onboard estimation scheme and control strategy.

We will introduce the definition of coordination frames and notations of parameters in Subsection 
  \ref{subsec:notation} and then introduce the control strategy in \ref{subsec:control} and estimation 
  framework in \ref{subsec:est}. The detailed estimation design is introduced in Section \ref{sec:est}.

\subsection{Problem Formulation and Notations}\label{subsec:notation}

\begin{figure}[h!t]
	\centering
	\input{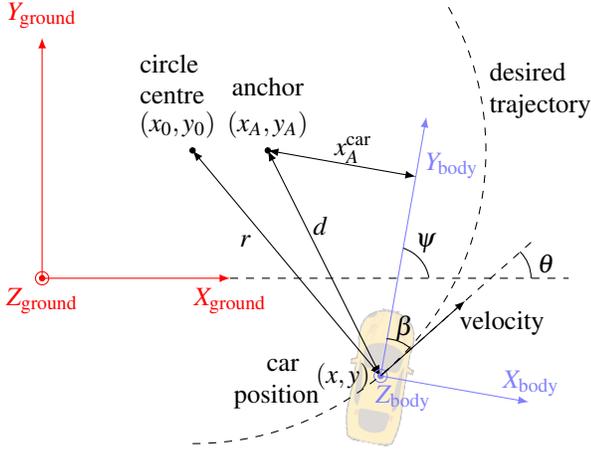}
	\caption{The top view of coordination frames and parameters when drifting.}
	\label{fig:car_top}
\end{figure}

The top view of our coordination frame definition and system layout is presented is Fig. \ref{fig:car_top}.

The ground frame (depicted as red in Fig. \ref{fig:car_top}) is a three dimensional Cartesian coordinate 
  with X-axis pointing to east, Y-axis pointing to north and Z-axis pointing straight up (the opposite 
  direction of gravity). The body frame is also a three dimensional Cartesian coordinate (depicted as blue 
  in Fig.~\ref{fig:car_top}) with X-axis pointing to the right of car, Y-axis pointing straight ahead of 
  the vehicle and Z-axis pointing straight up. Since the vehicle is moving on the ground plane with 
  negligible centroid height change, we concentrate on the state on XY plane and illustrate the top view 
  of the frames and parameters in Fig. \ref{fig:car_top}.

The position of vehicle centroid in ground frame is denoted as $(x,y,z)$ and their time derivatives 
  by $(\dot{x},\dot{y},\dot{z})$. The heading angle $\psi$ of the vehicle is the angle of the body Y-axis 
  projected into the ground frame XY-plane. The direction of vehicle velocity is the direction of 
  vector $(\dot{x},\dot{y})$, whose attitude angle is denoted as $\theta$. The angle between car heading 
  direction and velocity direction is denoted as $\beta$, and is refereed as the \textbf{sideslip angle}:
  \begin{equation}
    \beta\triangleq \theta-\psi.
  \end{equation}
  A negative sideslip angle ($0>\beta>-\pi$) represents that the vehicle is slipping to the right of 
  its heading direction as shown in Fig. \ref{fig:car_top}.

Our proposed estimation scheme aims at obtaining accurate state estimations 
  $\hat{x}, \hat{y}, \hat{\dot{x}}, \hat{\dot{y}},\hat{\psi}$ with a high frequency and admissible delays, 
  with only onboard cameras and IMU. The control strategy aims at maintaining a circular drifting state 
  with stable sideslip angle and circle center, i.e., the desired system states satisfy:

\begin{align}
x(t)&=x_0+r_{\rm ref}  \cos(2\pi\frac{t}{\tau}), \\
y(t)&=y_0+r_{\rm ref}  \sin(2\pi\frac{t}{\tau}), \\
\psi(t)&=2\pi\frac{t}{\tau} +\frac{\pi}{2}-\beta_{\rm ref},
\end{align}
where $\tau$ represents the time required for a drift cycle.
 
In order to accurately estimate the state and compensate for the floating error from inertial measurements 
  and image distortion error from cameras, we introduce a visual anchor to calibrate the position 
  estimation when driving. The visual anchor is an object easily identified by the camera and has known 
  position $(x_A,y_A)$. By estimating the relative distance $d$ to anchor and the X-axis deviation of 
  the anchor in body frame $x^{\rm car}_A$, one can obtain the car position $(x,y)$ in the ground frame.

\begin{figure*}[b!ht]
	\centering
	\tikzset{every picture/.style={scale=0.9}}
\usetikzlibrary{arrows}
\tikzset{rounded corners}
\begin{tikzpicture}
	\draw [-latex,fill=yellow!10!white] (5.5,4) rectangle (10.25,-1);
	\node at (6.5,1.5) {Vehicle};
	
	\draw [fill=blue!10!white] (8,3.5) rectangle node[text width=1.6cm, align=center]{IMU} (9.5,2.5) ;
	\draw [fill=blue!10!white] (7.5,0.5) rectangle node[text width=1.9cm, align=center]{ZED Stereo \\ Camera} (10,-0.5) ;
	\draw [fill=blue!10!white] (7.5,2) rectangle node[text width=2.9cm, align=center]{RealSense Depth Camera} (10,1) ;
	
	\node at (12,3.25) {$\psi$};
	\draw [-latex](9.5,3) -- (14.5,3);
	\draw [-latex](12.5,3) -- (12.5,2.25);
	
	\draw [-latex](10,1.5) -- (11.5,1.5);
	\node at (11,1.75) {$x^{\rm car}_A$};
	\node at (11,1.25) {$y^{\rm car}_A$};
	
	\draw [-latex, fill=green!10!white] (11.5,2.25) rectangle node[text width=2.1cm, align=center]{\footnotesize Position Calibration by Visual Anchor} (14,0.75) ;
	\draw [-latex](14,1.5) -- (14.5,1.5);
	
	\draw [-latex](10,0) -- (14.5,0);
	\node at (12.5,-0.25) {\footnotesize Cruel position};
	
	\draw [-latex, fill=green!10!white] (14.5,3.5) rectangle node[text width=2.0cm, align=center]
	{Asynchronous Kalman Estimator} (18,-0.5) ;

	\node [text width=1.0cm, align=center] at (15.1,0) {\scriptsize position $\hat{x}\ \hat{y}$};
	\node [text width=1.0cm, align=center] at (16.25,0) {\scriptsize  velocity $\hat{\dot{x}} \ \hat{\dot{y}}$};
	\node [text width=1.0cm, align=center] at (17.4,0) {\scriptsize yaw angle $\hat{\psi}$};
	
	\draw [dashed] (14.5,0.5) rectangle (15.75,-0.5);
	\draw [dashed] (15.75,0.5) rectangle (16.75,-0.5);
	\draw [dashed] (16.75,0.5) rectangle (18,-0.5);
	
	
	\draw (15.25,-0.5)--(15.75,-1.5);
	\draw (16.25,-0.5)--(15.75,-1.5);
	\draw (16.25,-0.5)--(16.75,-1.5);
	\draw (17.25,-0.5)--(16.75,-1.5);

	\draw [latex-](8,-3.5)--(8.5,-3.5);
	\draw [latex-](10,-3.5)--(15.75,-3.5);
	\draw [latex-](10,-5)--(16.75,-5);
	\draw (16.75,-5)--(16.75,-1.5);
	
	\node at (12.5,-1.5) {$\hat{x}\ \hat{y}\ \hat{\dot{x}} \ \hat{\dot{y}}$};
	\node at (12.5,-3) {$\hat{x}\ \hat{y}\ \hat{\dot{x}} \ \hat{\dot{y}}$};
	\node at (12.5,-4.5) {$\hat{\dot{x}} \ \hat{\dot{y}}\ \hat{\psi}$};
	
	\draw [-latex, fill=green!10!white] (5.5,-4.5) rectangle node [text width=2.5cm, align=center]
	{\footnotesize Resilient Slip Angle Estimator} (10,-5.5) ;
	
	\draw [-latex, fill=green!10!white] (8.5,-3) rectangle node [text width=1.5cm, align=center]
	{\footnotesize Trajtory Memory} (10,-4) ;
	\draw [-latex, fill=green!10!white] (5.5,-3) rectangle node [text width=2.5cm, align=center]
	{\footnotesize Resilient Radius Estimator} (8,-4) ;
	\draw [-latex] (6.75, -3) -- (6.75, -2.5);
	\node at (7.3,-2.75) {$x_0,y_0$};

	
	\draw (2.5,-3.5) -- (5.5,-3.5);
	\draw [-latex](2.5,-3.5) -- (2.5,-0.65);
	
	\draw [-latex, fill=red!10!white] (2.5,-0.5) ellipse (0.2 and 0.2);
	\draw [-latex](3.5,-0.5) -- (2.65,-0.5);
	\draw (3.5,-0.5) -- (3.5,-2);
	\draw (5.5,-2) -- (3.5,-2);
	\draw (15.75,-1.5) -- (15.75,-3.5);
	
	\draw [-latex](15.75,-2.0) -- (10,-2);
	
	\draw [-latex, fill=green!10!white] (5.5,-1.5) rectangle node [text width=2.5cm, align=center]
	{\footnotesize Circle Path Planner} (10,-2.5) ;

	\draw (5.5,-5) -- (1,-5);
	\draw [-latex](1,-5) -- (1,2.3);
	
	\node at (4.5,-1.75) {$r_{\rm ref}$};
	\node at (4.5,-3.25) {$\hat{r}$};
	\node at (4.5,-4.75) {$\hat{\beta}$};

	

	\draw [-latex](2.5,-0.25) -- (2.5,0.25);
	\draw [-latex, fill=red!10!white] (1.6,1.6) rectangle node[text width=1.6cm, align=center]{Circle Controller} (3.4,0.3);
	
	\draw [-latex](3.4,1) -- (4.1,1);
	\draw [-latex, fill=red!10!white] (4.25,1) ellipse (0.2 and 0.2);

	\draw [-latex](4.4,1) -- (5.5,1);
	
	\node at (1,3.75) {$\beta_{\mathrm{ref}}$};
	\node at (0.75,1.5) {$\beta$};
	
	\draw [-latex](3.4,2.5) -- (4.1,2.5);
	\draw [-latex, fill=red!10!white] (1.6,1.9) rectangle node[text width=1.6cm, align=center]{Sideslip Controller}(3.4,3.2);
	\draw [-latex](1.2,2.5) -- (1.6,2.5);
	\draw [-latex, fill=red!10!white] (1,2.5) ellipse (0.2 and 0.2);
	\draw [-latex](1,3.3) -- (1,2.7);
	
	\node at (4.25,1.0) {$+$};
	\node at (1,2.5) {$-$};
	\node at (2.5,-0.5){$-$};
	
	\draw [-latex](4.25,0.25) -- (4.25,0.8);
	
	\draw [-latex, fill=red!10!white] (4.25,2.5) ellipse (0.2 and 0.2);
	\draw [-latex](4.4,2.5) -- (5.5,2.5);
	\draw [-latex](4.25,3.5) -- (4.25,2.7);

	\node at (4.25,2.5) {$+$};

	\node at (4.25,3.75) {$\delta_{\mathrm{ff}}$};
	\node at (5.2,2.7) {$\delta$};
	\node at (4.25,0) {$\omega_{\mathrm{ff}}$};
	\node at (5.2,1.2) {$\omega$};
	
	\draw[dashed, red!30!black, thick]  (0,4.5) rectangle (5,-1);
	\node[text width=2.4cm, align=center, text=red!50!black] at (2.5,4) {Controller};

	
	\draw[dashed, green!30!black, thick]  (10.5,4.5) -- (18.5,4.5);
	\draw[dashed, green!30!black, thick]  (18.5,-5.75) -- (18.5,4.5);
	\draw[dashed, green!30!black, thick]  (5.25,-1.25) -- (5.25,-5.75);
	\draw[dashed, green!30!black, thick]  (18.5,-5.75) -- (5.25,-5.75);
	\draw[dashed, green!30!black, thick]  (10.5,4.5) -- (10.5,-1.25);
	\draw[dashed, green!30!black, thick]  (5.25,-1.25) -- (10.5,-1.25);
	
	\node[text=green!50!black] at (14,4) {On-board Estimation Scheme};
	
\end{tikzpicture}
	\caption{Overview of our control architecture.}
	\label{fig:control_scheme}
\end{figure*}
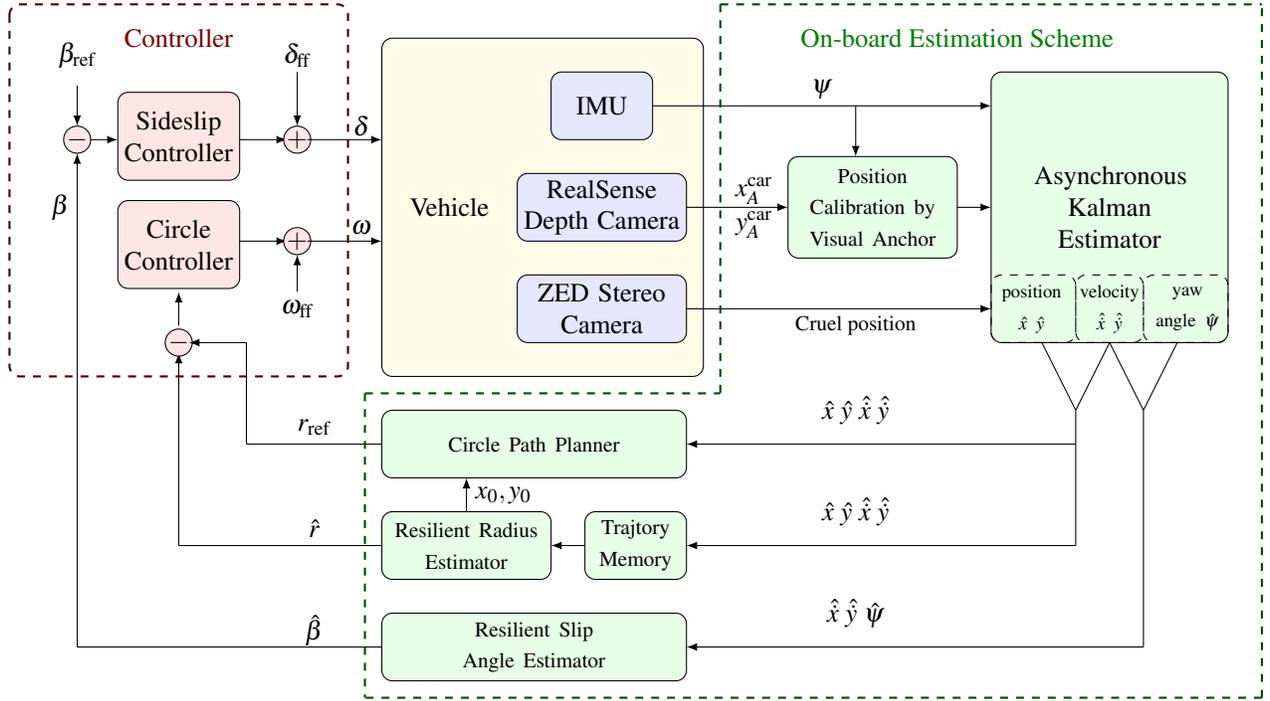

In the following two subsections, we propose our control strategy and estimation scheme for the circular 
  drifting task, the architecture of which is presented in Fig. \ref{fig:control_scheme}. Our race vehicle 
  is equipped with an IMU and two cameras which provide state measurements used for the estimator. 
  The system state are then utilized to calculate the slip angle $\beta$, trajectory radius $r$ and 
  expected radius $r_{\rm ref}$. These three critical features then feed into the controller.

\subsection{Two Loop Control Scheme}\label{subsec:control}
The basis of our controller is two feedback control loops (see Fig. \ref{fig:control_scheme}): 
  The sideslip controller stabilizes the sideslip angle $\beta$ by tuning the front wheel steering angle 
  $\delta$. The goal of this controller is to keep the vehicle in a roughly circular drifting trajectory, 
  by maintaining the sideslip angle $\beta$ as close as a constant reference angle $\beta_{\rm ref}$.
  The circle controller maintains a stable trajectory radius and circle center by controlling the 
  wheel rotational speed $\omega$.

We assume without loss of generality that the desired path is anticlockwise. The sideslip controller 
  and the circle controller are both PID controller with feed-forwards:
    \begin{align*}
    \delta(t)&=\delta_{\mathrm{ff}}+ K^{S}_Pe_\beta(t)+ K^{S}_I \int e_\beta(t) +K^{S}_D \cdot{e}_\beta(t), \\
    \omega(t)&=\omega_{\mathrm{ff}}+ K^{C}_Pe_r(t)+ K^{C}_I \int e_r(t) +K^{C}_D \cdot{e}_r(t),
    \end{align*}
  where $e_\beta(t)=\beta(t)-\beta_{\rm ref}, e_r(t)=r(t)-r_{\rm ref}$. $\beta_{\rm ref}$, $\delta_{\mathrm{ff}}$ 
  and $\omega_{\mathrm{ff}}$ are constant obtained from simulation and tire model identification
  . We introduce the calculation of $r_{\rm ref}$ in the following. 

\begin{figure}[h!]
      \centering
      \tikzset{every picture/.style={scale=0.9}}
      \begin{tikzpicture}

\draw[dashed,blue,line width=1pt]  (-3,1) ellipse (2 and 2); 

\filldraw  (-1,3) circle (1pt); 
\draw[-latex] (-1,3) -- (-2.5,3.625);

\draw [red,line width=1pt] (-1,3) arc (67.38:130:3.25);

\draw[line width=1pt] (-1,3) arc (67.38:20:4);
\node [text width=1.3cm] at (1,2.5) {real\\ trajectory};

\node at (-2.5,4) {$v$};

\node [blue,text width=1.3cm]  at (-6,0.5) {desired \\ trajectory};
\node[red,text width=1.3cm] at (-3.5,3.5)  {transition \\ trajectory};

\draw (-3,1)--(-1,3);
\draw[dashed] (-1,3)--(-0,4);

\draw (-0.5,3.5) arc (40:160:0.717);
\node at (-1,4) {$\phi$};  
\node at (-2.5,2) {$d$};

\filldraw [blue] (-3,1) circle (1pt); 
\node at (-3.65,0.75) {$(x_0,y_0)$};
\draw [-](-3,1)--(-4.75,2);
\node at (-3.8,1.7) {$r_0$};

\filldraw (-1,3) circle (1pt); 
\node at (-0.3,3.1) {$(x,y)$};
\end{tikzpicture}
      \caption{Illustration of the control of radius.}
      \label{fig:circle_control}
    \end{figure}
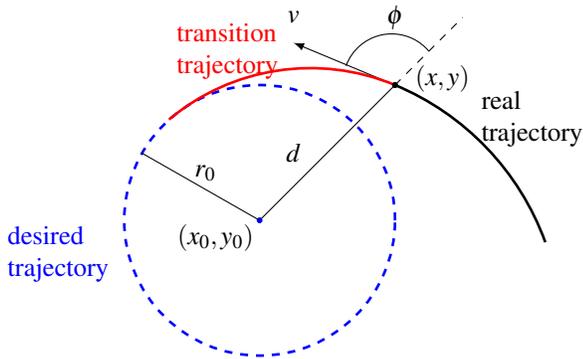

The design of the circle controller integrates both the radius control task and the circle center task by 
  keeping radius $r$ tracking the planned radius $r_{\rm ref}$. Circumnavigation using only bearing or 
  distance information has been well-explored in the literature (e.g. \cite{6705614}, \cite{ZHENG2015400} 
  and \cite{DONG2020108932}). We adopt a similar design idea and introduce the angle $\phi$ which is the 
  difference between the current velocity attitude angle $\theta=\arctan(\dot{y}/\dot{x})$ and the angle 
  of the drift center relative to the vehicle:
    \begin{equation*}
    \phi \triangleq \arctan(\dot{y}, \dot{x}) - \arctan{(y - y_{0}, x - x_{0})},
    \end{equation*} 
  where $(x_{0},y_{0})$ is the center of desired circular trajectory. We design the reference radius as
    \begin{equation}\label{eq:r_ref}
    r_{\rm ref} = r_{0}-\gamma(\pi/2-\phi),
    \end{equation} 
  where $\gamma > 0$ is the radius adjustment rate and $r_{0}$ is the desired radius.
  The design intuition is that the reference radius should adjust $\phi$ such that $\phi$ convergences 
  to $\pi/2$. It can be seen from equation \eqref{eq:r_ref} that when $\phi>\pi/2$, we have 
  $r_{\rm ref}>r_0$, and when $\phi<\pi/2$, we have $r_{\rm ref}<r_0$ which is illustrated in Fig.~\ref{fig:circle_control}.
  Moreover, by a change of coordinates, as is shown in Fig.~\ref{fig:circle_control}, the kinematics 
  of our vehicle can be characterized as $$\dot{d}=v \cos (\phi),\ \dot{\phi}=\frac{v}{r}-\frac{v}{d} \sin (\phi).$$
  We assume that the tracking of radius is perfect, i.e., $r=r_{\rm ref}$, and $v$ is constant, then 
  the dynamics of the vehicle will be 
    \begin{align}\label{eq:cir_sys}
    \dot{d}=v \cos (\phi),\ \dot{\phi}=\frac{v}{r_0-\gamma(\pi/2-\phi)}-\frac{v}{d} \sin (\phi).
    \end{align}
  By the same Lyapunov function as in Theorem 1 from \cite{drift_yang}, one can prove that the system in 
  \eqref{eq:cir_sys} has a set of globally asymptotically stable equilibrium: $$\phi=\pi/2+2k\pi, d=r_0, \ k\in\Zb.$$

\subsection{Overall Estimation Scheme}\label{subsec:est}

The sampling rate of different sensors usually varies greatly. Moreover, the sampling interval of sensors 
  may also be time-varying due to fluctuating processing time under environmental change. Therefore, it is 
  practical to develop an estimator that admits asynchronous and non-uniformly sampled measurements 
  and updates the estimates whenever there are new measurements.

Therefore, we suppose that the measurement time stamps constitute an ascending time sequence:
  \begin{equation*}
  0=t_0<t_1<t_2 <\cdots < t_k < \cdots .
  \end{equation*} 
Define the set of sampling time as 
  \begin{equation*}
  \Gamma \triangleq \{t_0,t_1,t_2,\cdots\} .
  \end{equation*}
We only concentrate on state estimation at time in the set $\Gamma$. Denote the time interval as
  \begin{equation}
    \Delta_k=t_{k}-t_{k-1}.
  \end{equation}
Since our goal is to maintain a stable circular drifting on a horizontal plane, we use a simplified set 
  of parameters to characterize the state of the vehicle. The system state is 
  $X(t)=\begin{bmatrix}x(t),y(t),v(t),\theta(t)\end{bmatrix}^\top$. We adopt a stable circular drifting 
  dynamic as following:
    \begin{align}
    x(t_{k+1})&=x(t_{k})+v(t_{k})\cdot\cos(\theta(t_{k}))\cdot\Delta_k,\\
    y(t_{k+1})&=y(t_{k})+v(t_{k})\cdot\sin(\theta(t_{k}))\cdot\Delta_k,\\
    v(t_{k+1})&=v(t_{k}), \\
    \theta(t_{k+1})&=\theta(t_{k})+v(t_{k})/r\cdot \Delta_k, 
    \end{align} 
where $v$ is the magnitude of the velocity and $r$ the desired radius. 
Similar dynamics are used by Zhu \textit{et al.} \cite{ZHU9305510} describing a stable circular movement for circular circumnavigation.

Moreover, the control input $\omega$ denotes the rotational speed of the wheels and changes 
  the velocity $v$. The control input $\delta$ is the front wheel steering angle and changes 
  the direction $\theta$ of velocity. Therefore, we obtain the following linear time-varying system:
    \begin{align}
      \begin{bmatrix}
        x(t_{k+1}) \\ y(t_{k+1}) \\ \theta(t_{k+1}) \\ v(t_{k+1})  
      \end{bmatrix}
      =
      \begin{bmatrix}
      1 &  0 & 0 & \cos(\theta(t_{k}))\Delta_k   \\
      0 &  1 & 0 & \sin(\theta(t_{k}))\Delta_k   \\
      0 &  0 & 1 & \Delta_k/r  \\
      0 &  0 & 0 & 1  
      \end{bmatrix}
      \begin{bmatrix}
      x(t_{k}) \\ y(t_{k}) \\ \theta(t_{k}) \\ v(t_{k})  \end{bmatrix} \notag \\
      +
      \begin{bmatrix}
      0& 0 \\
      0& 0 \\
      B_\delta& 0 \\
      0& B_\omega
      \end{bmatrix} 
      \begin{bmatrix}
      \delta(t_k) \\ \omega(t_k) 
      \end{bmatrix} .
    \end{align}
  We write it as a compact form and introduce the process noise $w_1(t)$:
    \begin{align}\label{eq:est_dynamic}
      X(t_{k+1})=A(t_{k}) X(t_{k})+Bu(t)+w_1(t),
    \end{align}
  where $w_1(t)$ is Gaussian process noise with zero mean and covariance matrix $Q(t)\succeq 0$. 

And the measurements from sensors at time $t_k\in\Gamma$ are:
  \begin{equation}\label{eq:y_i_def}
    Y(t_k)=C(t_k) X(t_k)+w_2(t_k), 
  \end{equation}
  where $Y$ is composed of position measurements from D435i camera $x_{\rm D435i},y_{\rm D435i}$, 
  position measurements from ZED camera $x_{\rm ZED}, y_{\rm ZED}$, and velocity angle obtained 
  from velocities and heading angle, i.e., 
    $$\theta_{\rm IMU}=\arctan(\dot{y}/\dot{x})+\psi_{\rm IMU},$$
  where $\psi_{\rm IMU}$ is the vehicle heading angle is obtained from IMU. The measurements and noise 
  are defined as:
  \begin{align}
    Y(t_k) \triangleq\begin{bmatrix}
      x_{\rm D435i}\\
      y_{\rm D435i}\\
      x_{\rm ZED}\\
      y_{\rm ZED}\\
      \theta_{\rm IMU}
    \end{bmatrix}  ,
    w_2(t_k) \triangleq\begin{bmatrix}
      w_{21}(t_k) \\
      \vdots \\
      w_{25}(t_k)
    \end{bmatrix}.
  \end{align}
  $C(t_k)$ is the corresponding coefficient whose elements are zeros when the corresponding measurement 
  is not available at time $t_k$ and are ones when they are available. $w_2(t_k) \in \mathbb{R}$ is 
  the measurement noise with covariance $R(t_k)\succeq 0$ independent of the noise process $\{w_1(t)\}$.

We introduce the following assumption to simplify the estimation update.
  \begin{assumption}
    At each sampling time $t_k\in\Gamma$, there is only one measurement update from one sensor.
  \end{assumption}
This assumption is with loss of generality since the value of time stamp can be accurate to the 
  nanosecond level and, in practice, no measurements have identical time stamp under this accuracy level.

In the following we establish the asynchronous extended Kalman filter.
  \begin{subequations}\label{eq:asy_kalman_pre}
    \begin{align}
      &\hspace{-20pt}\textbf{Prediction steps:} \notag \\
      &\hat{X}_{\m}(t_k) = A(t_{k-1})\hat{X}(t_{k-1}) + \Delta_kB u(t_{k-1}),\\
      &P_{\m}(t_{k}) = A(t_{k-1})P(t_{k-1})A^{\top}(t_{k-1}) + Q(t_{k-1}). 
    \end{align}
  \end{subequations}
Define $i(k)$ as the index of sensor that reports a new measurement at time $k$. Define $R_{i(k)}(t_{k})$ 
  as the $i(k)$-th diagonal element of matrix $R(t_{k})$.
  \begin{subequations}\label{eq:asy_kalman_up}
    \begin{align}
      & \textbf{Update steps:} \notag \\
      & K(t_{k}) = P_{\m}(t_{k}) C^{\top}_{i(k)}(t_{k}) \left(C_{i(k)}(t_{k}) P_{\m}(t_{k}) C^{\top}_{i(k)}(t_{k})+R_{i(k)}(t_{k}) \right)^{-1}  , \label{eq:def_Kk_asy} 	\\
      & P(t_{k}) = (I-K(t_{k}) C_{i(k)}(t_{k})) P_{\m}(t_{k}), \label{eq:def_Pt_asy} \\
      & \hat{X}(t_{k}) = \hat{X}_{\m}(t_{k}) + K(t_{k})  \left(y_{i(k)}(t_{k})-C_{i(k)}(t_{k}) \hat{X}_{\m}(t_{k}) \right) . \label{eq:def_xt_asy}
    \end{align}
  \end{subequations}

\begin{remark}\label{rk:Ki=0}
	Notice that $K(t_{k})$ is a $n\times 1$ matrix and the estimation update \eqref{eq:def_xt_asy} only 
    involves the measurement at sensor $i(k)$.
\end{remark}

\section{State Estimation by onboard Sensors}\label{sec:est}

\subsection{Sensor Configuration}

In our experiments, we use a 1/10 scale race car equipped with NVIDIA JETSON AGX XAVIER as the 
  computing unit which can complete 2 different state update procedures at high frequency: 
  state from ZED camera update at 100Hz and state from D435i camera update at 60Hz. At the same time,  
  our visual state estimation and visual position tracking module operate at 60 Hz. 

Our sensor modality is image-based with a stereo camera ZED 2 installed back of the car and
  an RGB-D camera REALSENSE D435i installed in the front of this car. 
  The stereo camera is intended to provide a ground frame state estimation whereas the RGB-D camera is 
  used for providing an accurate ground frame position.

A motion capture system broadcasts pose information at 120Hz, which is further used by a Kalman filter
  to provide an accurate ground frame state as the reference.
  
\begin{figure}[h!t]
  \centering
  \input{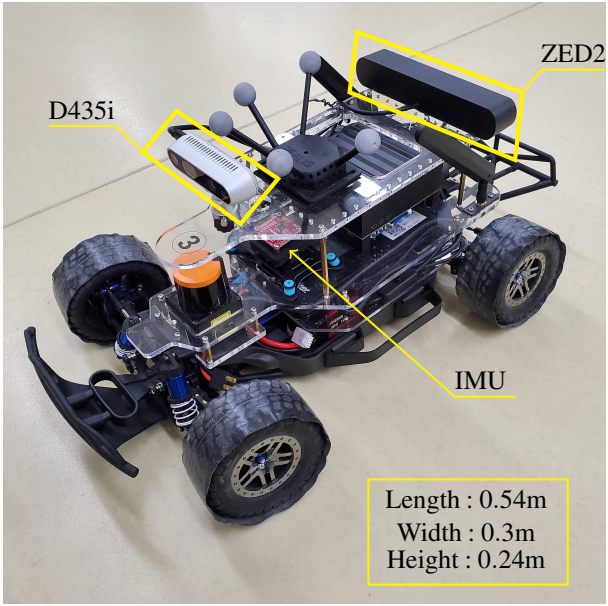}
  \caption{The model racecar platform and sensor configuration.}
  \label{fig:car}
\end{figure}

\subsection{Primary position estimation by ZED stereo camera}

The ZED uses visual tracking of its surroundings to understand the movement of the user or system 
  holding it. As the camera moves in the real world, it reports its new position and orientation. 
  Pose information is outputted at the frame rate of the camera, which is 100Hz in our situation.

ZED camera initialized automatically when it is started, and generate a three-dimensional 
  coordinate system based on its current position and attitude, in which camera is located at origin 
  and yaw angle is 0. Although we do not care about the deviation of the initial position, 
  the angular deviation between the ZED frame and the ground frame will affect the calculation of 
  the sideslip angle. Therefore, we need to collect the yaw angle $\psi_{\rm 0}$ in the ground frame when 
  ZED camera initializing, calculate the rotation matrix $T_{\rm 0}$ according to $\psi_{\rm 0}$, 
  correct ZED frame to ground frame. Since the installation position of the ZED camera is not on the 
  center point of the vehicle, it is also necessary to refer to the yaw angle $\psi$ of the vehicle for 
  coordinate transformation with rotation matrix $T$ when calculating the actual position of the vehicle.

  \begin{equation}
    \begin{bmatrix} x_{\rm car} & y_{\rm car} \end{bmatrix} = 
      \begin{bmatrix} x_{\rm car}^{\rm ZED} & y_{\rm car}^{\rm ZED} \end{bmatrix} \cdot T_{\rm 0} + p_{\rm X} \cdot T,
  \end{equation} 
  where
  \begin{equation*}
    T_{\rm 0}  = 
    \begin{bmatrix}
      \cos{(\psi_{\rm 0} )} & \sin{(\psi_{\rm 0} )} \\ 
      -\sin{(\psi_{\rm 0} )} & \cos{(\psi_{\rm 0} )}
    \end{bmatrix},
  \end{equation*}
  \begin{equation*}
    T = 
    \begin{bmatrix}
      \cos{(\psi)} & \sin{(\psi)} \\ 
      -\sin{(\psi)} & \cos{(\psi)}
    \end{bmatrix},
  \end{equation*} 
  and $p_{\rm X} = \begin{bmatrix}p_{\rm x} & p_{\rm y}\end{bmatrix}$ means 
  the difference between the installation position of ZED camera and the center point of vehicle under the 
  ground frame.

However, the cumulative error of ZED camera will gradually affect the accuracy of fusion. 
  To fix this error, we introduce another camera into our structure.

\subsection{Position Online Calibration by Visual Anchor}

In order to deal with the cumulative error caused by ZED, we use the pure vision method and depth information 
  to estimate the position of our vehicle. This method has many application cases, for example, Han \textit{et al.} 
  \cite{han2016vehicle} using a mono-camera to estimate the distance between different vehicles, and 
  Diaz-Cabrera \textit{et al.} \cite{diaz2015robust} using a robust technique to detect traffic lights during 
  both day and night conditions and estimate their distance. In our case, we apply this method through 
  a special anchor placed in the environment, a pure blue barrel. In order to avoid collisions, 
  the anchor is suspended in the air at a certain height. At this height, the vehicle can pass under 
  the anchor, and the camera can capture the necessary features of the anchor.

Color features are widely used in object detection. 
  In this paper, we first filter the colors in the image to extract the same or similar colors as 
  the anchor. After that, the extracted content is further filtered through the actual width height ratio 
  of the anchor. We then obtain the coordinates of anchor $(x_{\rm A}^{\rm img}, y_{\rm A}^{\rm img})$ 
  in the image, corresponding to $(x_{\rm A}^{\rm car}, y_{\rm A}^{\rm car})$ in the car frame.

  \begin{figure}[h!]
    \centering
    \input{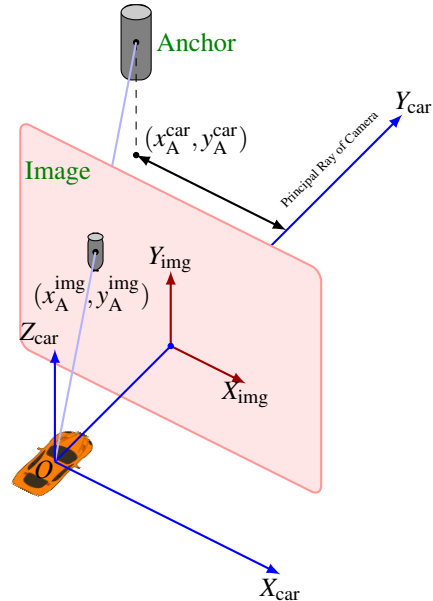}
    \caption{The visual anchor and relative values from the car frame.}
    \label{fig:car_barrel}
  \end{figure}

Based on the principle of camera imaging and similarity, we know that the included angle $\theta$ 
  between the anchor and the optical axis of the camera is proportional to the transverse distance between 
  the center of the projection of anchor on image and the center of the image $x_{\rm A}^{\rm img}$. 
  At the same time, the area occupied by the anchor in image $S_{\rm img}$ is inversely proportional 
  to the square of the distance between anchor and camera. Since the shape of the projection of anchor 
  on the image is similar to a rectangle, We conclude that the distance $d$ between anchor and camera is 
  inversely proportional to the height $h_{\rm img}$ (or width $w_{\rm img}$) of the projection of 
  anchor on image:

  \begin{subequations}\label{eq:distance}
    \begin{align}
      & d = k_1/{\sqrt{S_{\rm img}}}, \label{eq:dis} \\
      & \theta = k_2 x_{\rm A}^{\rm img} . \label{eq:theta}
    \end{align}
  \end{subequations}

From data, we can fit the specific values of $k_1$ and $k_2$. Assuming that the width height ratio of 
  anchor is fixed as $k$, then we have
  \begin{equation}
    d = k_1/{\sqrt{S_{\rm img}}} = k_1/{\sqrt{k w_{\rm img}^2}} = k_1'/w_{\rm img},
  \end{equation} 
  where
  \begin{equation*}
    k_1' = k_1/\sqrt{k}.
  \end{equation*} 

D435i can provide RGB image and depth data at the same time. Because the field angle of view of the 
  depth camera is different from that of the RGB camera, we need to align the two images before 
  obtaining depth information, but the calculation resources required for alignment will greatly reduce 
  our update frequency. Therefore, we get the approximate correspondence between the two images 
  based on data. Using the coordinate of anchor in the RGB image, and the approximate correspondence, 
  we directly obtained the distance between the bucket and the car using depth information. 
  Then we fuse the two distances to reduce the error.

Through the $d$ and $\theta$ obtained previously, we can calculate the relative coordinate between anchor
 and vehicle in the vehicle coordinate system.

\begin{equation}\label{eq:relative coordinate}
  p_{\rm X} = \begin{bmatrix} x_{\rm A}^{\rm car} & y_{\rm A}^{\rm car} \end{bmatrix} 
    - \begin{bmatrix} x_{\rm c}^{\rm car} & y_{\rm c}^{\rm car} \end{bmatrix} = d \cdot \begin{bmatrix}\sin{(\theta)} & \cos{(\theta)}\end{bmatrix}.
\end{equation}

Then calculate the rotation matrix $T$ through yaw angle $\psi$ (provided by IMU), so that the 
  relative coordinate relationship in the vehicle coordinate system can be transformed into the ground frame. 
  \begin{equation}\label{eq:global car coordinate}
    \begin{bmatrix} x_{\rm c} & y_{\rm c} \end{bmatrix} = \begin{bmatrix} x_{\rm A} & y_{\rm A} \end{bmatrix} + p_{\rm X} \cdot T,
  \end{equation}
where $T$ is the rotation matrix mentioned before, and the $p_{\rm X}$ here is the difference between the 
  installation position of D435i camera and the center point of vehicle under the ground frame.
  $\begin{bmatrix} x_{\rm A} & y_{\rm A} \end{bmatrix}$, $\begin{bmatrix} x_{\rm c} & y_{\rm c} \end{bmatrix}$
  represent coordinate of anchor and car under the ground frame respectively.

So far, we have a more accurate coordinate compared to the ZED camera. However, the maximum update frequency 
  of this coordinate is only 60Hz, which cannot meet the control requirements, so we need to 
  fusion ZED camera and D435i camera to meet the requirements.

\subsection{Subsequent Processing of State Estimation}\label{esti}

In this subsection, we introduce the detailed processing procedure of estimated states. 
  We intend to obtain three critical values used for drifting control, i.e., 
  (1) Slip Angel $\beta$, which is the angle between vehicle attitude and velocity direction, 
  as shown in Fig. \ref{fig:car_top}; (2) Trajectory radius $r$; (3) Reference radius $r_{\rm ref}$, 
  used to guide the vehicle moving to the desired circle center.

\textbf{(1) Resilient Slip Angle Estimation}

According to the definition of slip angle $\beta$, one can obtain
\begin{align}
	\beta= \arctan\left({\dot{y}}/{\dot{x}}\right)-\psi.
\end{align}

Since the estimation of $\dot{x},\dot{y}$ are noisy and may contain sparse outliers, we obtain the following resilient estimator:
\begin{align*}
\hat{\beta}(t_k)=
\arctan \left(\frac{\hat{\dot{y}}(t_k)}{\hat{\dot{x}}(t_k)}\right)-\hat{\psi}(t_k) ,
\end{align*}
if the following inequality holds
\begin{align}\label{eq:beta_th}
\left|\arctan \left(\frac{\hat{\dot{y}}(t_k)}{\hat{\dot{x}}(t_k)}\right)-\arctan \left(\frac{\hat{\dot{y}}(t_{k-1})}{\hat{\dot{x}}(t_{k-1})}\right)\right|<h\cdot \Delta_k,
\end{align}
where $h$ is the slip angle abrupt change threshold.
If \eqref{eq:beta_th} is  violated, the velocity angle is predicted using a circular drifting dynamic:
\begin{align*}
\hat{\beta}(t_k)=
\arctan \left(\frac{\hat{\dot{y}}(t_{k-1})}{\hat{\dot{x}}(t_{k-1})}\right)+\frac{\hat{v}(t_{k-1})}{\hat{r}(t_k)}\cdot \Delta_k -\hat{\psi}(t_k),
\end{align*}
where $\hat{r}$ is obtained from the radius estimator (will be introduced in the following) and $\hat{v}$ is obtained from the Kalman filter.

\textbf{(2) Resilient Radius Estimation}

In order to maintain a circle drifting with desired radius and center, we need to accurately estimate 
  the parameters of the circle that best fits into the current trajectory. The problem of circle fitting has 
  been well-investigated since the early 50's originated from microwave engineering. An efficient method of 
  estimating the radius of a 2D trajectory is the well-known KASA algorithm \cite{KASA}. It is simple and 
  efficient since it only involves solving a $3\times 3$ linear equality regardless of the number of 
  data points. However, it is not resilient to sparse outliers, which often occurs when using onboard 
  sensors. Main efforts aiming at resilient circle fitting focus on optimization stability 
  \cite{robust_circle14} and noise tolerance~\cite{circle_robot}. In the following, we propose a novel 
  resilient circle fitting algorithm insensitive to sparse outliers. 

Suppose that we have a sequence of $N$ points from time $k-N+1$ to $k$ on 2D plane: $\{x(t),y(t)\}_{t=k-N+1}^{k}$. 
  A classic least-square fitting algorithm proposed by Delonge and adopted by KASA\cite{KASA} solves the 
  following optimization problem:
  \begin{subequations}
    \begin{align}
    \underset{x_0,\ y_0,\ r,\ \{r_t\}}{\rm minimize}&\quad \sum_{t=k-N+1}^{k} \left(r_t^2-r^2\right)^2 \\
    \text{s.t.}& \quad (x(t)-x_0)^2+(y(t)-y_0)^2=r_t^2, \\
    &\quad  t=k-N+1,\cdots,k. \notag
    \end{align}
  \end{subequations}
where the solution of $(x_0,y_0)$ is the coordination of the circle center and $r$ is the radius. 
  Since few of the data points may be corrupted by large bias due to image detection failures or extra 
  computation delay, we propose the following resilient radius estimation scheme:
  \begin{subequations}
    \begin{align}
    \underset{x_0,y_0,r,\{r_t\},\{a_t\}}{\rm minimize}&\quad \sum_{t=k-N+1}^{k} \left(r^2_t-r^2\right)^2 +\lambda \sum_{t=k-N+1}^k |a_t|\\
    \text{s.t.}& \quad (x(t)-x_0)^2+(y(t)-y_0)^2+a_t=r_t^2, \\
    &\quad t=k-N+1,\cdots,k. \notag
    \end{align}
  \end{subequations}
  where $\lambda>0$ is a weighting parameter. By introducing the $\ell_1$ regularization term, 
  the solution tolerates sparse outlier points $(x(t),y(t))$ by tuning $a_t$. The following 
  Fig. \ref{fig:circle_fit} illustrates the data fitting results with normal points (the left figure) 
  and with partly corrupted points (the right figure).

  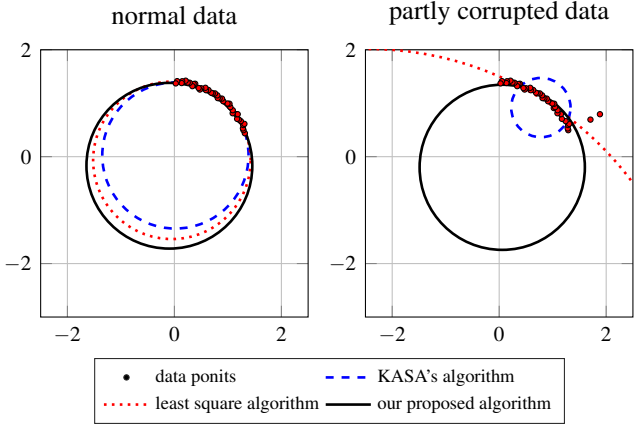
\begin{figure}[ht!]
  	\centering
  	\begin{tikzpicture}
\begin{axis}
[	width=1.4in,
	height=1.4in,
	at={(0in,0in)},
	scale only axis,
xmin=-2.5,
xmax=2.5,
ymin=-3,
ymax=2,
	title={normal data},
	xticklabel style = {font=\footnotesize},
	yticklabel style = {font=\footnotesize},
	axis background/.style={fill=white},
	xmajorgrids,
	ymajorgrids,
	legend columns=2, 
	legend style={at={(0.2,-0.4)}, anchor=south west, legend cell align=left, align=left, draw=white!15!black, font=\scriptsize}]
	
	 \addplot [scatter,only marks,scatter/use mapped color={fill=red},mark size=1.0pt]
	table[row sep={\\}]
	{\\
 1.31695  0.438211\\
1.29153  0.495296\\
1.28114  0.528689\\
1.28807  0.591908\\
1.30889  0.613455\\
1.24189  0.664643\\
1.20623  0.69341\\
1.17255  0.707302\\
 1.15439  0.795177\\
1.18347  0.794302\\
1.0894   0.812338\\
1.09144  0.868753\\
1.04591   0.926538\\
1.02975   0.990549\\
0.993186  0.994385\\
 0.913429  1.03749\\
0.890262  1.08594\\
0.845472  1.09054\\
0.810598  1.09581\\
0.802965  1.14723\\
0.760609  1.18292\\
0.71729   1.18491\\
0.688754  1.19627\\
 0.636218  1.24694\\
0.591107  1.24367\\
0.580389  1.28852\\
0.502455  1.27997\\
0.475983  1.25609\\
0.453809  1.28383\\
0.389016  1.36384\\
0.359735  1.35976\\
 0.336149   1.3196\\
0.286739   1.38282\\
0.248569   1.38608\\
0.210623   1.42145\\
0.145579   1.41073\\
0.135016   1.3774\\
0.0516843  1.41288\\
0.0255824  1.37071\\
	};\addlegendentry{data ponits}

\addplot[blue,line width=1pt, dashed] {10 1};
\addplot[red,line width=1pt, dotted] {10 1};
\addplot[black,line width=1pt] {10 1};
\legend{data ponits,KASA's algorithm,least square algorithm,our proposed algorithm};

\draw [blue,line width=1pt, dashed] (0.018991,0.0225373) circle (1.36705031); 
\draw [red,line width=1pt, dotted] (-0.0470914,-0.0684829) circle (1.4722); 
\draw [black,line width=1pt] (-0.0935361481,-0.1695123) circle (1.5501355); 

\end{axis}

\begin{axis}
[	width=1.4in,
height=1.4in,
at={(1.7in,0in)},
scale only axis,
xmin=-2.5,
xmax=2.5,
ymin=-3,
ymax=2,
	title={partly corrupted data},
xticklabel style = {font=\footnotesize},
yticklabel style = {font=\footnotesize},
axis background/.style={fill=white},
xmajorgrids,
ymajorgrids,
legend style={at={(0.01,0.01)}, anchor=south west, legend cell align=left, align=left, draw=white!15!black, font=\scriptsize},
]

\addplot [scatter,only marks,scatter/use mapped color={fill=red},mark size=1.0pt]
table[row sep={\\}]
{\\
	1.29153  0.495296\\
	1.28114  0.528689\\
	1.28807  0.591908\\
	1.30889  0.613455\\
	1.24189  0.664643\\
	1.70623  0.69341\\
	1.17255  0.707302\\
	1.15439   0.795177\\
	1.88347   0.794302\\
	1.0894    0.812338\\
	1.09144   0.868753\\
	1.02091   0.909914\\
	1.04591   0.926538\\
	1.02975   0.990549\\
	0.993186  0.994385\\
	0.913429  1.03749\\
	0.890262  1.08594\\
	0.845472  1.09054\\
	0.810598  1.09581\\
	0.802965  1.14723\\
	0.760609  1.18292\\
	0.71729   1.18491\\
	0.688754  1.19627\\
	0.636218  1.24694\\
	0.591107  1.24367\\
	0.580389  1.28852\\
	0.502455  1.27997\\
	0.475983  1.25609\\
	0.453809  1.28383\\
	0.389016  1.36384\\
	0.359735  1.35976\\
	0.336149   1.3196\\
	0.286739   1.38282\\
	0.248569   1.38608\\
	0.210623   1.42145\\
	0.145579   1.41073\\
	0.135016   1.3774\\
	0.0516843  1.41288\\
	0.0255824  1.37071\\
} ;

\draw [blue,line width=1pt, dashed] (0.77934947,0.918528) circle (0.5542671);
\draw [red,line width=1pt, dotted] (-2.473077,-4.16984) circle (6.18588523); 
\draw [black,line width=1pt] (0.0571249,-0.1987259) circle (1.5446233); 

\end{axis}
\end{tikzpicture}
  	\caption{Circle fitting results of different algorithms.}
  	\label{fig:circle_fit}
  \end{figure}


\section{Experiment}

We test the algorithm on our 1/10 scale race car. In the beginning, we use the motion capture system to provide the vehicle state to ensure that 
  the vehicle can enter the drift state stably. When the vehicle can drift around the anchor stably 
  after a certain time, the sensing data source is switched to onboard sensors.

\begin{figure}[h!t]
  \centering
\begin{tikzpicture}

\definecolor{color0}{rgb}{1,0.498039215686275,0.0549019607843137}
\definecolor{color1}{rgb}{0.172549019607843,0.627450980392157,0.172549019607843}

\begin{axis}[
tick align=outside,
tick pos=left,
x grid style={white!69.0196078431373!black},
xmin=0.5, xmax=4.5,
xtick={1,2,3,4},
xticklabels={Truth,ZED,D435i,EKF},
xtick style={color=black},
y grid style={white!69.0196078431373!black},
ylabel={sideslip angle error (radian)},
ymin=-0.00234153741942332, ymax=0.0491722858078898,
ytick style={color=black}
]

\addplot [blue]
table {%
0.775 0.00486343819098334
1.225 0.00486343819098334
1.225 0.0160729820776915
0.775 0.0160729820776915
0.775 0.00486343819098334
};
\addplot [purple]
table {%
1.775 0.00681478897080079
2.225 0.00681478897080079
2.225 0.0228342208006369
1.775 0.0228342208006369
1.775 0.00681478897080079
};
\addplot [red]
table {%
2.775 0.00400038374681949
3.225 0.00400038374681949
3.225 0.0155562773139288
2.775 0.0155562773139288
2.775 0.00400038374681949
};
\addplot [green]
table {%
3.775 0.00661694918280542
4.225 0.00661694918280542
4.225 0.0151358425344266
3.775 0.0151358425344266
3.775 0.00661694918280542
};

\addplot [black]
table {%
1 0.00486343819098334
1 1.92284851319169e-05
};
\addplot [black]
table {%
1 0.0160729820776915
1 0.0328758647970127
};
\addplot [black]
table {%
0.8875 1.92284851319169e-05
1.1125 1.92284851319169e-05
};
\addplot [black]
table {%
0.8875 0.0328758647970127
1.1125 0.0328758647970127
};

\addplot [black]
table {%
2 0.00681478897080079
2 4.56215626662981e-05
};
\addplot [black]
table {%
2 0.0228342208006369
2 0.0468307483884665
};
\addplot [black]
table {%
1.8875 4.56215626662981e-05
2.1125 4.56215626662981e-05
};
\addplot [black]
table {%
1.8875 0.0468307483884665
2.1125 0.0468307483884665
};

\addplot [black]
table {%
3 0.00400038374681949
3 0
};
\addplot [black]
table {%
3 0.0155562773139288
3 0.032886117291409
};
\addplot [black]
table {%
2.8875 0
3.1125 0
};
\addplot [black]
table {%
2.8875 0.032886117291409
3.1125 0.032886117291409
};

\addplot [black]
table {%
4 0.00661694918280542
4 2.45456744107919e-06
};
\addplot [black]
table {%
4 0.0151358425344266
4 0.0278928594400818
};
\addplot [black]
table {%
3.8875 2.45456744107919e-06
4.1125 2.45456744107919e-06
};
\addplot [black]
table {%
3.8875 0.0278928594400818
4.1125 0.0278928594400818
};

\addplot [color0]
table {%
0.775 0.0101746569404453
1.225 0.0101746569404453
};
\addplot [color1, mark=triangle*, mark size=3, mark options={solid}, only marks]
table {%
1 0.0109608731497298
};
\addplot [color0]
table {%
1.775 0.0134814424141747
2.225 0.0134814424141747
};
\addplot [color1, mark=triangle*, mark size=3, mark options={solid}, only marks]
table {%
2 0.016132598339139
};
\addplot [color0]
table {%
2.775 0.00857375936561622
3.225 0.00857375936561622
};
\addplot [color1, mark=triangle*, mark size=3, mark options={solid}, only marks]
table {%
3 0.0107328835750843
};
\addplot [color0]
table {%
3.775 0.0109607976310822
4.225 0.0109607976310822
};
\addplot [color1, mark=triangle*, mark size=3, mark options={solid}, only marks]
table {%
4 0.0113299193178211
};
\end{axis}

\end{tikzpicture}
  \caption{Absolute value of sideslip angle error in 11 circles of drifting.}
  \label{fig:drift_angle}
\end{figure}
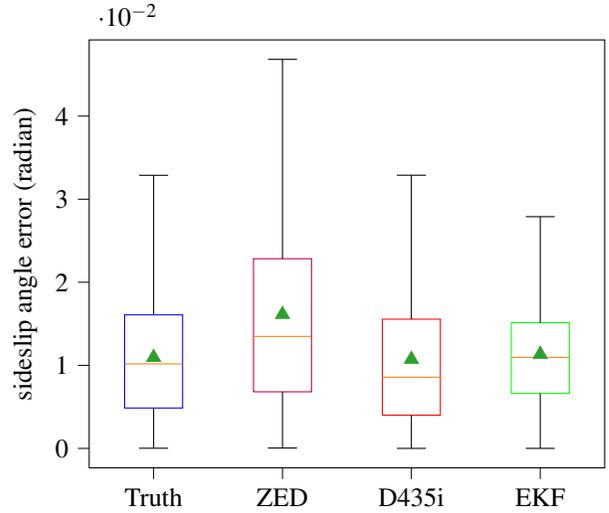

Our race car finished 11 circles of drifting in 50 seconds. The reference radius is 1 meter and sideslip angle -1.4 radians. The drifting control and estimation performance are shown in Fig. \ref{fig:drift_radius} and \ref{fig:drift_angle} by absolute value of radius error and sideslip angle error, respectively. The data labelled by ``Truth" is provided by the motion capture system. On the contrary, the data lablled by ``ZED", ``D435i", and ``EKF" are the estimation error by using merely ZED, D435i, and these two, respectively. 
  

\begin{figure}[h!t]
  \centering
\begin{tikzpicture}

\definecolor{color0}{rgb}{1,0.498039215686275,0.0549019607843137}
\definecolor{color1}{rgb}{0.172549019607843,0.627450980392157,0.172549019607843}

\begin{axis}[
tick align=outside,
tick pos=left,
x grid style={white!69.0196078431373!black},
xmin=0.5, xmax=4.5,
xtick={1,2,3,4},
xticklabels={Truth,ZED,D435i,EKF},
xtick style={color=black},
y grid style={white!69.0196078431373!black},
ylabel={radius error (m)},
ymin=-0.053414397871133, ymax=1.12170235529379,
ytick style={color=black}
]

\addplot [blue]
table {%
0.775 0.0929529728154996
1.225 0.0929529728154996
1.225 0.222591338674269
0.775 0.222591338674269
0.775 0.0929529728154996
};
\addplot [purple]
table {%
1.775 0.102409580541415
2.225 0.102409580541415
2.225 0.490480785749877
1.775 0.490480785749877
1.775 0.102409580541415
};
\addplot [red]
table {%
2.775 0.0417903855621609
3.225 0.0417903855621609
3.225 0.166757445020804
2.775 0.166757445020804
2.775 0.0417903855621609
};
\addplot [green]
table {%
3.775 0.0505872865057083
4.225 0.0505872865057083
4.225 0.158126855535842
3.775 0.158126855535842
3.775 0.0505872865057083
};

\addplot [black]
table {%
1 0.0929529728154996
1 2.08470231863878e-05
};
\addplot [black]
table {%
1 0.222591338674269
1 0.412584821557047
};
\addplot [black]
table {%
0.8875 2.08470231863878e-05
1.1125 2.08470231863878e-05
};
\addplot [black]
table {%
0.8875 0.412584821557047
1.1125 0.412584821557047
};

\addplot [black]
table {%
2 0.102409580541415
2 3.81812282279625e-05
};
\addplot [black]
table {%
2 0.490480785749877
2 1.06828795742266
};
\addplot [black]
table {%
1.8875 3.81812282279625e-05
2.1125 3.81812282279625e-05
};
\addplot [black]
table {%
1.8875 1.06828795742266
2.1125 1.06828795742266
};

\addplot [black]
table {%
3 0.0417903855621609
3 0
};
\addplot [black]
table {%
3 0.166757445020804
3 0.352079312812797
};
\addplot [black]
table {%
2.8875 0
3.1125 0
};
\addplot [black]
table {%
2.8875 0.352079312812797
3.1125 0.352079312812797
};

\addplot [black]
table {%
4 0.0505872865057083
4 0.000349424834066081
};
\addplot [black]
table {%
4 0.158126855535842
4 0.280227815655351
};
\addplot [black]
table {%
3.8875 0.000349424834066081
4.1125 0.000349424834066081
};
\addplot [black]
table {%
3.8875 0.280227815655351
4.1125 0.280227815655351
};

\addplot [color0]
table {%
0.775 0.156926132176186
1.225 0.156926132176186
};
\addplot [color1, mark=triangle*, mark size=3, mark options={solid}, only marks]
table {%
1 0.163669350382152
};
\addplot [color0]
table {%
1.775 0.226593852685318
2.225 0.226593852685318
};
\addplot [color1, mark=triangle*, mark size=3, mark options={solid}, only marks]
table {%
2 0.341356060462102
};
\addplot [color0]
table {%
2.775 0.0934741453393514
3.225 0.0934741453393514
};
\addplot [color1, mark=triangle*, mark size=3, mark options={solid}, only marks]
table {%
3 0.111363974016623
};
\addplot [color0]
table {%
3.775 0.102096344716995
4.225 0.102096344716995
};
\addplot [color1, mark=triangle*, mark size=3, mark options={solid}, only marks]
table {%
4 0.105729445204763
};
\end{axis}

\end{tikzpicture}
  \caption{Aboslute value of drift radius error in 11 circles of drifting. The bottom and top of the box represent the first and third quartiles, and the band inside the box represents the median of the data. The ends of the whiskers represent the minimum and maximum of the data. The triangle stands for the average value.}
  \label{fig:drift_radius}
\end{figure}
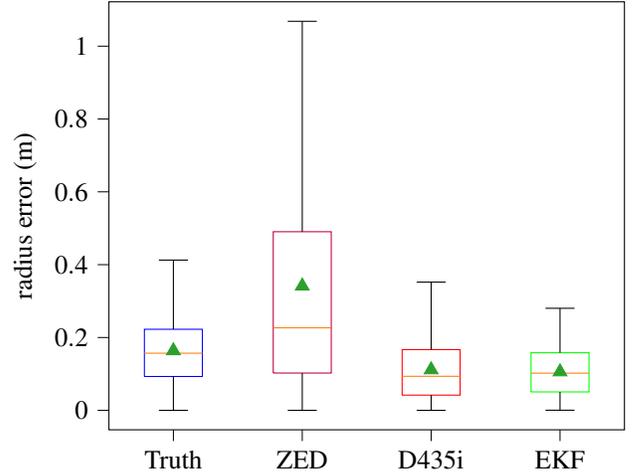

 It can be seen from the``Truth" data that our vehicle can track the given radius and sideslip angle quite well. Although the estimation performance by combining ZED and D435i is not significantly better than that of D435i, the frequency of D435i is only 60Hz, which is far lower than the 100Hz required for drifting control. The data frequency of ZED can meet the demand, but the estimation error of ZED is too large to make the vehicle drift stably for a long time. Our fusion algorithm strikes a good balance between these two, which not only improves the data frequency but also ensures accuracy.

\section{Conclusion}

In this paper, we implement our method on a 1/10 scale race car, it can realize stable drift with 
  a given center and radius. We propose a state estimation algorithm 
  using onboard cameras and IMU for aggressive racecar drifting control. The accuracy of our algorithm 
  when the anchor is in the field of view of D435i is comparable to that of the motion capture system. 
  At the same time, by combining with the tracking algorithm of the ZED camera, the update frequency of 
  vehicle status can reach a maximum of 160Hz, which can ensure the normal operation of the control 
  algorithm at the frequency of 100Hz. 
  In future works, we intend to solve the problem of missing anchor by increasing the number of anchors 
  or cameras, or other sensors to identify and locate anchors. In addition, the error caused by the ZED 
  camera might be solved by using a more accurate positioning algorithm.

\bibliographystyle{IEEEtran}
\bibliography{ref.bib} 

\end{document}